\documentclass[10pt, a4paper, conference, compsocconf]{IEEEtran}
\ifCLASSINFOpdf
  \usepackage[pdftex]{graphicx}
  
\else
\fi
\usepackage[cmex10]{amsmath}
\usepackage[tight,footnotesize]{subfigure}
\begin{document}
\title{Constrained Neural Style Transfer for Decorated Logo Generation}


\author{\IEEEauthorblockN{
 Gantugs Atarsaikhan,
 Brian Kenji Iwana,
 Seiichi Uchida}
\IEEEauthorblockA{Graduate School of Information Science and Electrical Engineering\\
 Kyushu University,
 Fukuoka, Japan\\ Email: \{gantugs.atarsaikhan, brian, uchida\}@human.ait.kyushu-u.ac.jp}
 
}


%Telephone: (800) 555--1212, Fax: (888) 555--1212}


\maketitle

\begin{abstract}
Making decorated logos requires image editing skills, without sufficient skills, it could be a time-consuming task.
While there are many on-line web services to make new logos, they have limited designs and duplicates can be made. 
We propose using neural style transfer with clip art and text for the creation of new and genuine logos.
We introduce a new loss function based on distance transform of the input image, which allows the preservation of the silhouettes of text and objects.
% The proposed method constrains style transfer to only a designated area.
The proposed method constrains style transfer only around the designated area.
We demonstrate the characteristics of proposed method.
Finally, we show the results of logo generation with various input images.
%We will explain the proposed method in details and provide some experimental results.
\end{abstract}


% \fi
\IEEEpeerreviewmaketitle

\section{Introduction}

% Background
There is a difficulty of designing logos with decorations.
If one does not possess image editing skills, a lot of time and energy will be wasted for making logos.
Although, there are on-line tools that can generate logos easily.
These websites use a heuristically mutating choice selector, which the user has to select options to generate the logo.
An example of this is Logomaster\footnote{https://logomaster.ai/}.
However, the problem with those kinds of websites is that the number of designs is limited, and there is a possibility for a duplicate.

\begin{figure}[!t]
\centerline{
\subfigure[Content Image]{\includegraphics[width=0.3\columnwidth]{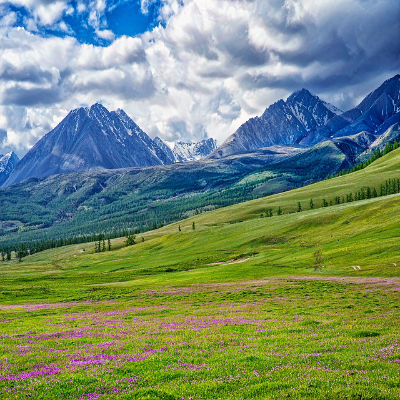}%
\label{landscape}}
\hfil
\subfigure[Style Image]{\includegraphics[width=0.3\columnwidth]{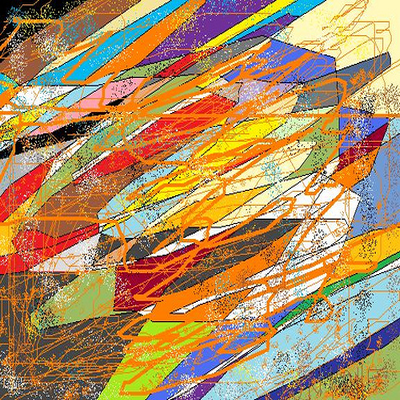}%
\label{modernart}}
\hfil
\subfigure[Generated Image]{\includegraphics[width=0.3\columnwidth]{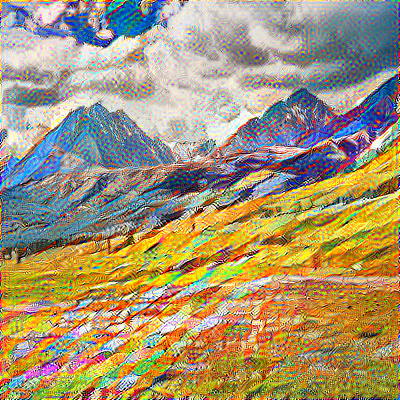}%
\label{land_art}}
}
\caption{An example of neural style transfer. Styles such as, textures and local details of the style image~\ref{modernart} has been transferred onto the main structure of content image~\ref{landscape} , resulting in the generated image~\ref{land_art}.}
\label{fig_sim}
\end{figure}

In recent years, style transfer using convolutional neural networks has been an active field.
There have been an abounding number of works for style transfer between two images to generate new images. 
Gatys et al.~\cite{Gatys} introduced the neural style transfer algorithm.
Neural style transfer uses Convolutional Neural Network~(CNN)~\cite{lecun1998gradient} to generate images by synthesizing a content image and a style image.
In neural style transfer, local features of the style image transfers onto the structure of a content image.
An example of neural style transfer is shown in Fig.~\ref{fig_sim}.
Also, ConvDeconv style network is used to transfer styles of image in real time by Johnson et al.~\cite{Johnson}.
Moreover, style transfer has been achieved by using Generative Adversarial Networks~(GAN) by Isola et al.~\cite{Pix2Pix}.
They successfully used GANs to map input images to newly generated images.

\begin{figure}
\centerline{
\subfigure[Text as a content image and its decorated logo.]{\includegraphics[width=0.45\columnwidth]{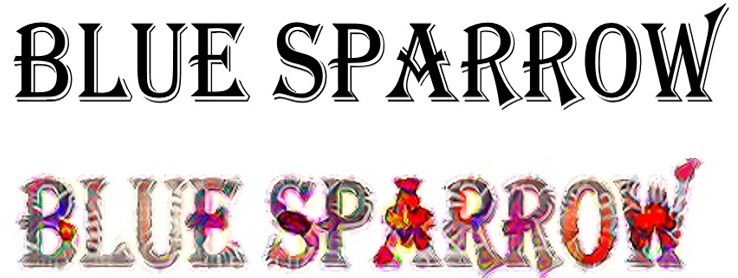}
\label{animal_shape}}
\hfill
\subfigure[Simple clip art as a content image and its decorated logo.]{\includegraphics[width=0.45\columnwidth]{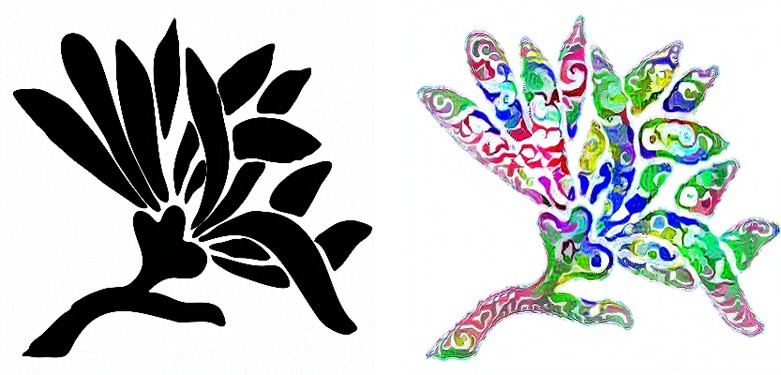}
\label{img_logo_ex}}
}
\caption{Examples of content images and a logo which content images will be transformed into.}
\label{fig_silhouette}
\end{figure}

% Purpose
The purpose of this paper is to propose a novel method of generating decorated logos automatically.
The term decorated logo refers to a decorated text (word-mark) or a symbol with decorated text.
In the proposed method, any pattern images can be used as style image.
But, the type of content image is fixed as clip art or binary silhouette like images as shown in examples of Fig.~\ref{fig_silhouette}.
In order to generate logos more clearly, we propose a new loss function in addition to the already existing ones in neural style transfer.
%We will explain the new loss function with some experimental results.

% Main contribution
The main contributions are summarized as follows.
\begin{enumerate}
\item The introduction of a new loss function, which is designed for maintaining the shape as possible as it could.
\item A method to generate new and genuine logos easily.
\end{enumerate}

% Remaining of the paper
The remaining of this paper is organized as follows.
Section~\ref{related} describes related work regarding in neural style transfer and logo generation.
Section~\ref{nst} describes neural style transfer algorithm and its mechanism.
Section~\ref{new_loss} explains about the new loss function, which is implemented into neural style transfer algorithm.
Section~\ref{results} shows experimental results in logo generation.
Finally, we will conclude our work and discuss about future work in Section~\ref{conclusion}.

% \begin{figure}
% \centerline{
% \subfigure[Example of a content image]{\includegraphics[width=0.45\columnwidth]{fig/compete.jpg}
% \label{animal_shape}}
% \hfill
% \subfigure[Example of a logo]{\includegraphics[width=0.45\columnwidth]{fig/compete_m_flower.jpg}
% \label{img_logo_ex}}
% }
% \caption{Examples of content images and a logo which content images will be transformed into.}
% \label{fig_silhouette}
% \end{figure}

% \begin{figure}
% \centering
% \includegraphics[width=0.3\columnwidth]{fig/compete_m_flower.jpg}
% \caption{Example of a logo.}
% \label{img_logo_ex}
% \end{figure}

% \begin{figure}
% \centering
% \includegraphics[width=0.3\columnwidth]{fig/silhouettes.jpg}
% \caption{Examples of content image.}
% \label{animal_shape}
% \end{figure}

\begin{figure*}%[!t]
\centering
\includegraphics[width=1.5\columnwidth]{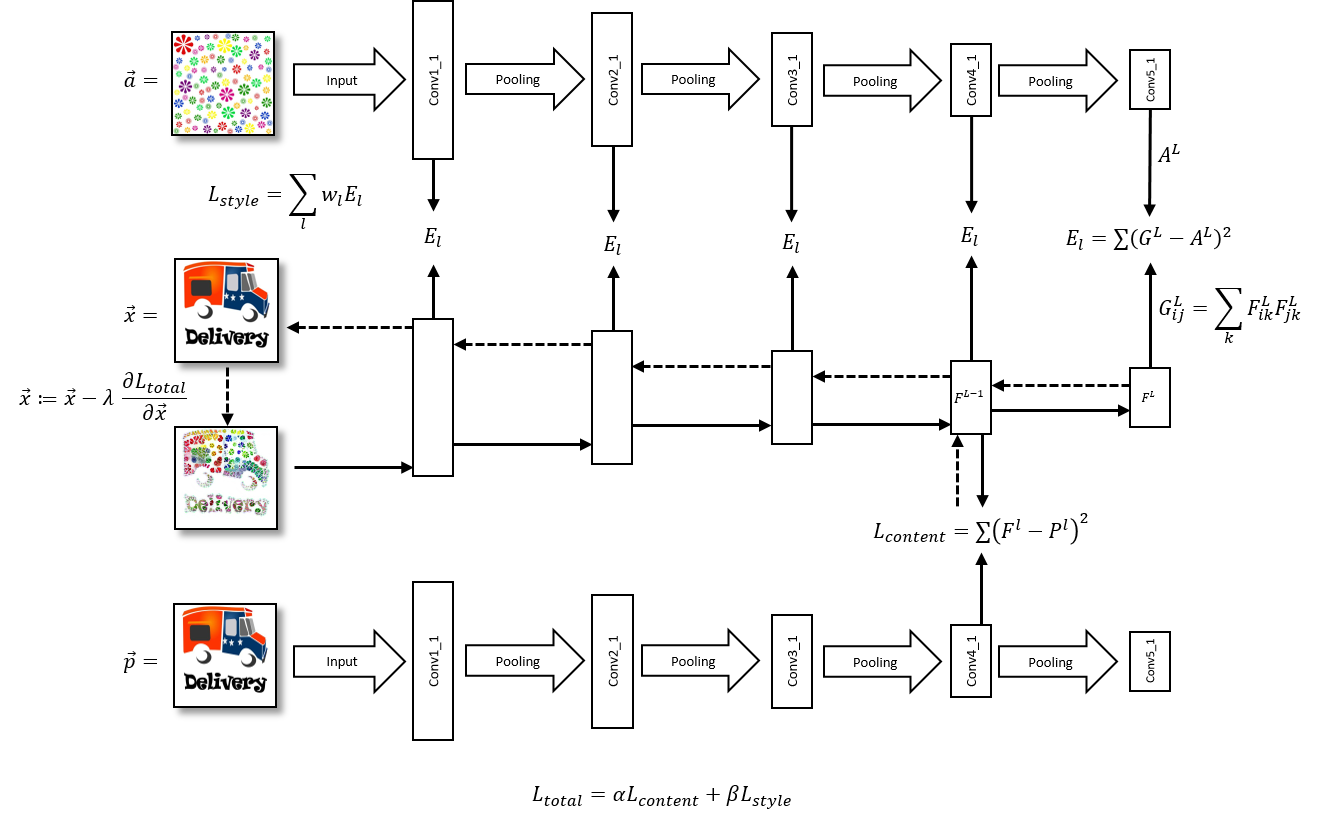}
\caption{The process flow of neural style transfer.}
\label{NST}
\end{figure*}

\section{Related Work}
\label{related}

One solution for logo generation is the use of genetic algorithms. 
Logo designing website Mark Maker\footnote{https://emblemmatic.org/markmaker/} uses a user-selected word to generate first generation of recommended logos.
Then depending on selections of the user, the system produces next generation of logos.
This process happens until the user finds their desired logo.

There have been few attempts to generate fonts automatically.
% Suveeranont et al.~\cite{Rapee} generated fonts from user handwriting sample, 
Tsuchiya et al.~\cite{Tsuchiya2014} used example fonts to determine predictive features.
Also, works has been done to generate fonts using interpolation of fonts~\cite{campbell2014learning,Uchida2015}.
Lately, a method to generate fonts using neural style transfer has been proposed~\cite{tugs}.

In the field of neural style transfer~\cite{Gatys}, there have been many methods introduced on transferring styles of art to an image~\cite{jing2017neural}.
Li and Wand~\cite{li2016precomputed} used Markovian GANs for better texture synthesis.
Chen et al.~\cite{chen2016fast} and Ulyanov et al.~\cite{ulyanov2016texture} increased calculation speed of neural style transfer and Gatys et al.~\cite{gatys2016preserving} preserved features of the content image.
Because of that, neural style transfer is used on video~\cite{anderson2016deepmovie} as well as sparse images~\cite{champandard2016semantic}.
The main advantage of neural style transfer is simplicity. There is no need of handcrafted features or heuristic rules.

\section{Neural Style Transfer}
\label{nst}

The basic principle of neural style transfer~\cite{Gatys} is to extract content representations and style representations of input images, and mix them into new image using a pre-trained CNN.
As a CNN, we used the Visual Geometry Group Network~(VGGNet)~\cite{Simonyan}.
The VGGNet was trained for image recognition with ImageNet dataset.
Because of its deep neural network design, the VGGnet is suitable for extracting content and style representations of an input image.

\subsection{Content and Style Representations}

With a given input image to the VGGNet, filter responses to every layer is produces as feature map. 
Feature maps on selected layers can be considered as the content representation of an input image.
A content representation on lower layers are more similar to the input image, where as a content representation on higher layers loses global features of the input image.

In order to obtain the style representation of an input image, a feature space, which is designed to capture texture information is used. 
This feature space can be built on any layers of the CNN. 
It consists of feature correlations given by the Gram matrix $G^l$ in multiple layers. 
The Gram matrix $G^l$ is given as,
\begin{equation}
     G_{ij}^l = \sum_{k}F_{ik}^lF_{jk}^l,
\end{equation}
where $F^l_{ik}$ and $F^l_{jk}$ refer to feature maps $i$ and $j$ in layer $l$.
The reason to use multiple layers is to obtain a consistent and multi-scale representation of the input image, thereby capturing only its texture information.

\subsection{Neural Style Transfer}

Fig.~\ref{NST} shows the process of transferring the style from a style image $\vec{a}$ onto a content image $\vec{p}$.
First, a content image $\vec{p}$ inputs to the VGGNet and its feature maps in selected layer are stored as the content representation $P^l$ on $l^{th}$ layer. 
Next, a style image $\vec{a}$ passes through the network. The sum of Gram matrices on every layer are computed and stored as style representation $A^L$ of a style image.

Then, the image to be generated $\vec{x}$, which is initialized as the content image, passes through the network. Using its feature maps, the content representation $F^l$ and the style representation $G^L$ of the generated image are computed on same layers as the respective representations.

With content and style representations, loss functions used for generating images can be calculated. 
The content loss $L_\mathrm{content}$ is calculated as a sum of square difference between content representations of the content image $\vec{p}$ and the generated image $\vec{x}$ in the selected layer as shown in Eq.~\eqref{eq_content_loss}.
\begin{equation}
\label{eq_content_loss}
     L_\mathrm{content}(\vec{p},\vec{x},\vec{l}) = \frac{1}{2}\sum_{ij}(F_{ij}^l - P_{ij}^l)^2.
\end{equation}
Style loss $L_\mathrm{style}$ can be calculated as a sum of square differences between style representations of the style image $\vec{a}$ and the generated image $\vec{x}$ in every layer.
\begin{equation}
\label{eq_style_loss}
     L_\mathrm{style}(\vec{a},\vec{x}) = \sum_{l=0}^{L}w_lE_l,
\end{equation}
where
\begin{equation}
\label{eq_E_l}
    E_l = \frac{1}{4N_{l}^2M_{l}^2}\sum_{ij}(G_{ij}^l-A_{ij}^l)^2.
\end{equation}
In Eqs.~\eqref{eq_style_loss} and~\eqref{eq_E_l}, $w_l$ are weighting factors for the contribution of layer $l$ to style loss, $N_l$ is the number of filters in layer $l$, $M_l$ is the dimensions of layer $l$ .

Then, we can combine $L_\mathrm{content}$ and $L_\mathrm{style}$ into the total loss $L_\mathrm{total}$ by taking linear addition. 
Given $\alpha$ and $\beta$ are weighting factors for content and style representations respectively, ${L_{total}}$ is defined as, 
\begin{equation}
	L_{total} = \alpha L_\mathrm{content} + \beta L_\mathrm{style}.
\end{equation}
Lastly, the generated image $\vec{x}$ is gradually optimized to minimize the total loss $L_\mathrm{total}$, thus generating a new image that has contents of the content image and the style of the style image.

\section{Distance transform loss}
\label{new_loss}

With neural style transfer, the style from the style image is applied to the entire content image.
For the purpose of art, it is desirable to synthesize a whole image with aspects of the content image and style image.
However, logos often do not fit in basic primitive shapes.
Therefore, when style transferring onto the silhouette image of a logo, the style unnecessarily transfers to the background. 
Because of that, there is too much noise outside of the desired contents.

However, we cannot simply cut the shape from the generated image.
Because part of a shape which is generated by neural style transfer may have cut, causing it to look unnatural.
In order to produce more natural looking generated image, we propose a new loss function using distance transform of the input images.
The distance transfer loss constrains style transfer to the shape of a content image and its near vicinity.

% One advantage working with silhouette like images is that those images are already segmented or easy to segment as background and foreground.
% This advantage makes it easier to determining the distance transform image with almost no processing cost.
% A style image can be any image that has patterns, but it is more preferable to use an image with same background color as the content image.

\subsection{Distance Transform}

\begin{figure}[!t]
\centerline{
\subfigure[Silhouette]{\includegraphics[height=1.4in]{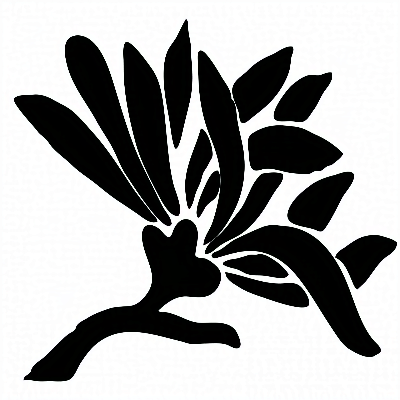}%
\label{silhouette_of_flower}}
\hfil
\subfigure[Distance transform]{\includegraphics[height=1.4in]{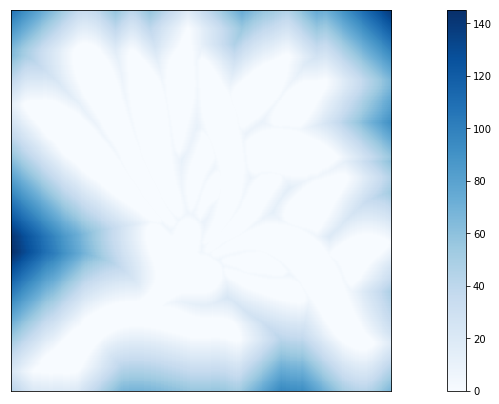}%
\label{distance_transform_image}}
}
\caption{Original binary image and its distance transform image (visualized by heat map). Inner side of the silhouette shapes are completely white but outer side gradually becomes darker with the increasing distance from the silhouette shape.}
\label{img_distance_transfer}
\end{figure}

\begin{figure}%[!t]
  \centerline{
    \subfigure[Content Image]{\includegraphics[width=0.3\columnwidth]{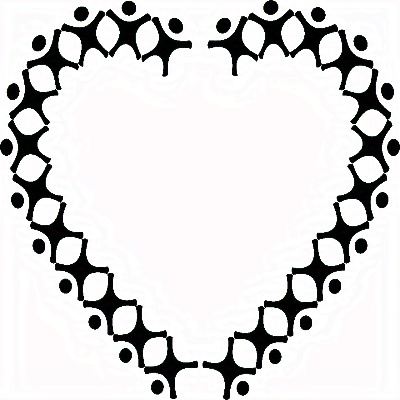}%
    \label{img_heart}}
    \hfil
    \subfigure[Style Image]{\includegraphics[width=0.3\columnwidth]{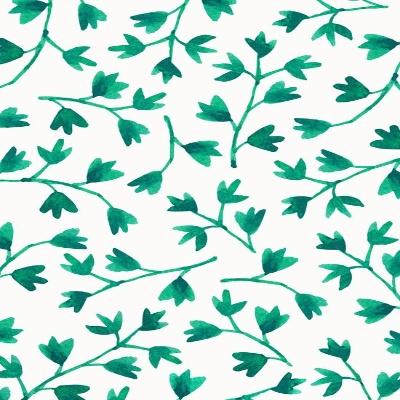}%
    \label{img_green_leaf}}
  }
  \centerline{
    \subfigure[Normal neural style transfer]{\includegraphics[width=0.3\columnwidth]{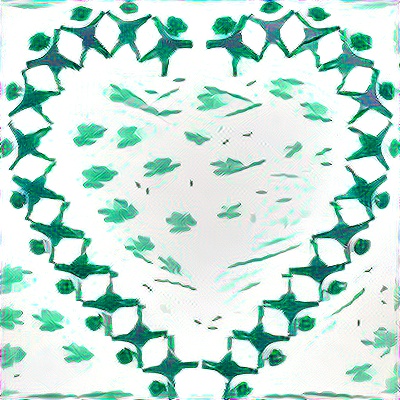}%
    \label{without}}
    \hfil
    \subfigure[With distance transform loss]{\includegraphics[width=0.3\columnwidth]{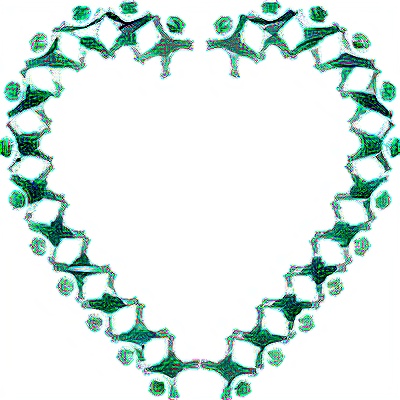}%
    \label{with}}
  }
  \caption{Comparison between using distance transform loss or not using.}
  \label{img_comparison}
\end{figure}

For each pixel in a binary image, the distance transform assigns a value, which is the distance to the nearest pixel that is silhouette. 
With distance transform, the value of the silhouette pixels become zero, and further the pixels are located from the silhouette higher the values become as shown in Fig.~\ref{img_distance_transfer}.

The distance transform image ${D}$ has same dimensions as an original image, but its pixel values are that of values of distance transform.
In this paper, we used Euclidean distance metric for its compatibility to different shapes in order to calculate the distance transform image ${D}$.

Once the distance transform image ${D}$ is computed, it can be manipulated to emphasize the specific group of pixels.
One way is to take pixel-wise power of distance transform image ${D}$, with power of two or more.
The values of pixels that are far from the silhouette shapes become large, whereas, values of nearby pixels do not change much.
For every pixel ${d_{ij}}$ of the distance transform image ${D}$, emphasis with power ${n}$ would look like,
\begin{equation}
\label{eq_emphasis}
d_{ij} =
  \begin{cases}
    0       & \quad \text{if inside of a silhouette} \\
    d_{ij}^n     & \quad \text{otherwise}
  \end{cases}
\end{equation}
However, ${n}$ should be two or higher.
In other words, we put weights on the pixels that are far from the silhouette shapes, which are more likely to be background pixel.
% During style transfer, due to its large value, the penalty for changing a pixel that is far from the silhouette shapes is much higher, and there will be no or little change in those pixels.
% On the other hand, the pixels that are closer to the shapes have much lower penalty due to its small values.
% Thus, the regions around the silhouette shape can be transformed.
% Also, pixels inside of silhouette shape have zero penalty, so most of the transformations will take place inside of shapes and a small region outside of shapes.

\subsection{Distance Transform Loss}

Given a content image $\vec{p}$, the generated image $\vec{x}$, the distance transform image ${D_\mathrm{content}}$ of the content image, and a natural number ${n}$, the distance transform loss ${L_\mathrm{distance}}$ would be, 
\begin{equation}
\label{eq_new_loss}
	L_\mathrm{distance} = \frac{1}{2}(\vec{p} \circ D^n_\mathrm{content} - \vec{x} \circ D^n_\mathrm{content})^2.
\end{equation}
In other words, first, pixel-wise multiplication of the content image $\vec{p}$ and its distance transform image $D^n_\mathrm{content}$ is calculated and stored.
Then, pixel-wise multiplication of generated image $\vec{x}$ and the distance transform image $D^n_\mathrm{content}$ of the content image is calculated.
Finally, the distance transform loss ${L_\mathrm{distance}}$ is calculated as, squared error between those multiplications.

Equation~\eqref{eq_new_loss} ensures that the penalty on pixels around the silhouette shapes are always smaller than those of further pixels.
While ${L_\mathrm{distance}}$ can preserve the shape of the silhouette and surroundings, it can also remove the noises from outside of the silhouettes shapes.
This process is emphasized by ${n}$, larger the ${n}$ the more noise will be removed.

We can implement the distance transform loss ${L_\mathrm{distance}}$ to neural style transfer by simply adding to the total loss ${L_\mathrm{total}}$ with weighting factor $\gamma$:
\begin{equation}
\label{new_total_loss}
	L_{total} = \alpha L_{content} + \beta L_{style} + \gamma L_{distance}.
\end{equation}

As shown in Fig.~\ref{img_comparison}, the distance transform loss constrains style transferring around the shape.
It also does not interfere with neural style transfer within the that area.

\begin{figure}%[!t]
  \centerline{
    \subfigure[Content Image]{\includegraphics[width=0.3\columnwidth]{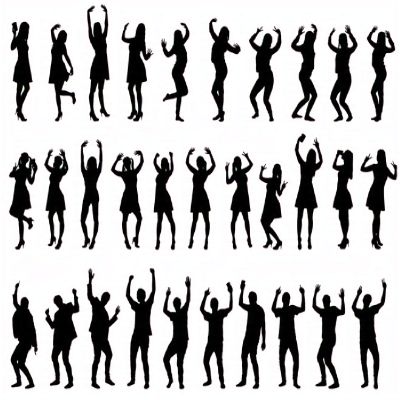}%
    \label{humans1}}
    \hfil
    \subfigure[Style Image]{\includegraphics[width=0.3\columnwidth]{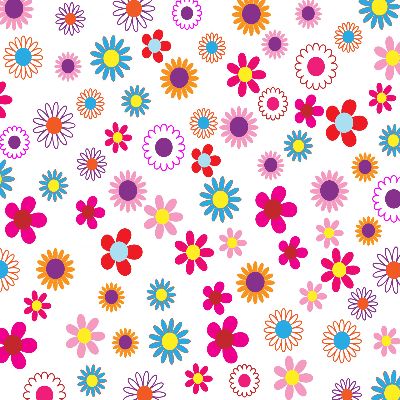}%
    \label{colorful_flower1}}
  }
  \centerline{
    \subfigure[${\alpha / \beta = 0.001}$]{\includegraphics[width=0.3\columnwidth]{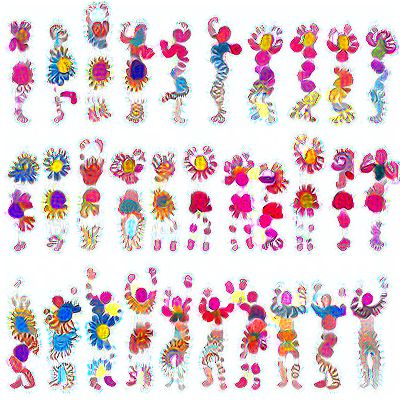}%
    \label{alpha001}}
    \hfil
    \subfigure[${\alpha / \beta = 0.01}$]{\includegraphics[width=0.3\columnwidth]{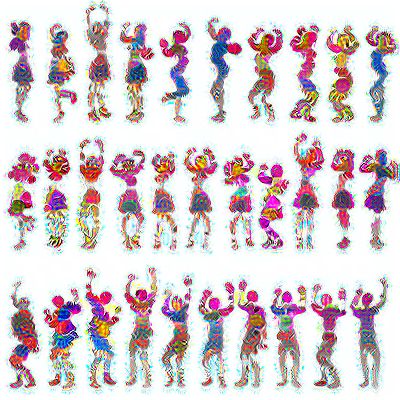}%
    \label{alpha01}}
    \hfil
    \subfigure[${\alpha / \beta = 1.0}$]{\includegraphics[width=0.3\columnwidth]{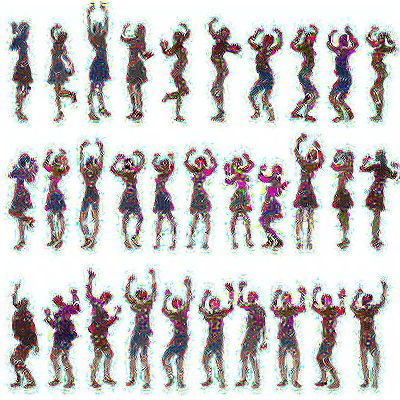}%
    \label{alpha10}}
  }
  \caption{Different ratio of weighting factors $\alpha / \beta$ with same $\gamma = 0.01$. With different $\alpha / \beta$ ratios, distance transform loss constrains the style transfer around shapes of silhouette indiscriminatingly.}
  \label{alpha}
\end{figure}

\section{Experimental Results}
\label{results}

The key contribution of this paper is to constrain the neural style transfer around the pre-segmented region using distance transform loss.
In this section, we will expose some aspects of neural style transfer constrained by distance transform loss.
Then, we will show some results of logo generation using proposed method.

Content representations were taken from layer "conv4\_2" of VGGNet.
Style representations were taken from the layers "conv1\_1", "conv2\_1", "conv3\_1", "conv4\_1", "conv5\_1".
Weighting factors for content image and style image, $\alpha$ and $\beta$ are constant at 0.001 and 1.0 respectively except for Fig.~\ref{alpha}.

\begin{figure}%[!t]
  \centerline{
    \subfigure[Content Image]{\includegraphics[width=0.3\columnwidth]{fig/humans1.jpg}%
    \label{humans}}
    \hfil
    \subfigure[Style Image]{\includegraphics[width=0.3\columnwidth]{fig/colorful_flower.jpg}%
    \label{colorful_flower}}
  }
  \centerline{
    \subfigure[${\gamma = 0.0001}$]{\includegraphics[width=0.3\columnwidth]{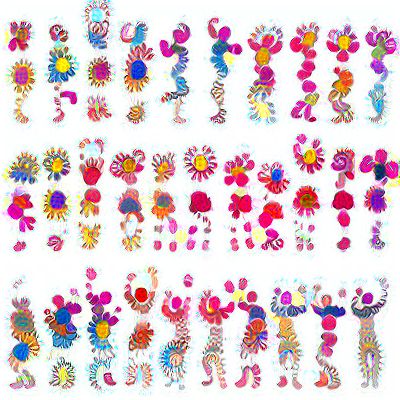}%
    \label{gamma0001}}
    \hfil
    \subfigure[${\gamma = 0.01}$]{\includegraphics[width=0.3\columnwidth]{fig/gamma01.jpg}%
    \label{gamma01}}
    \hfil
    \subfigure[${\gamma = 1.0}$]{\includegraphics[width=0.3\columnwidth]{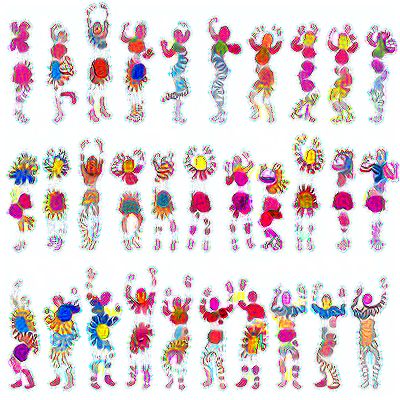}%
    \label{gamma10}}
  }
  \caption{Different values of weighting factor $\gamma$. When $\gamma$ is used, neural style transfer is more tightly constrained to silhouette shapes.}
  \label{gamma}
\end{figure}

\begin{figure}[!t]
  \centerline{
    \subfigure[Content Image]{\includegraphics[width=0.3\columnwidth]{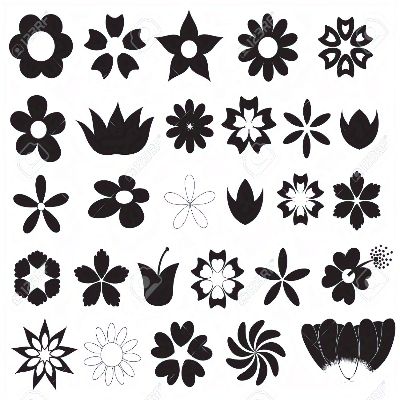}%
    \label{flowers}}
    \hfil
    \subfigure[Style Image]{\includegraphics[width=0.3\columnwidth]{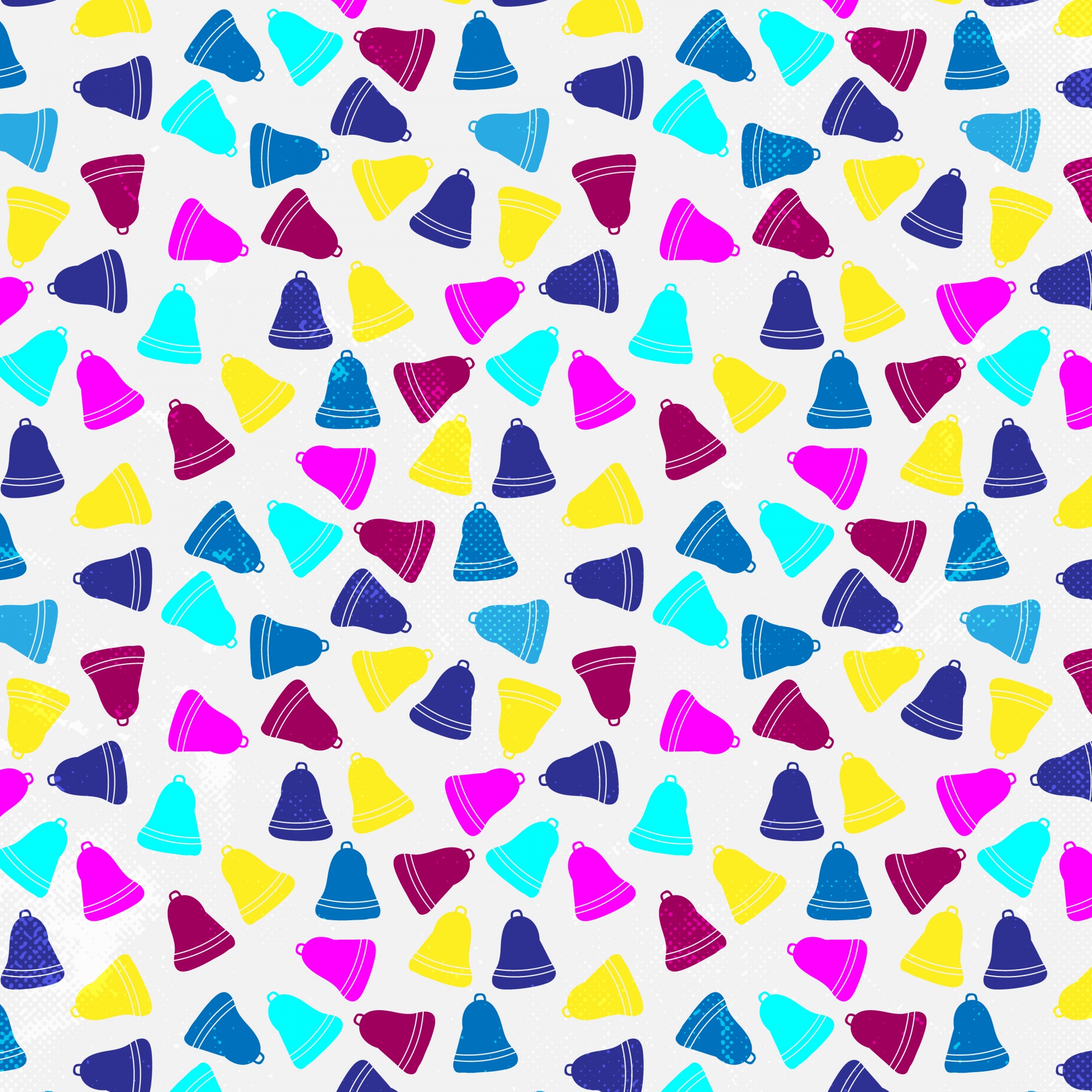}%
    \label{bells}}
  }
  \centerline{
    \subfigure[${D^1}$]{\includegraphics[width=0.3\columnwidth]{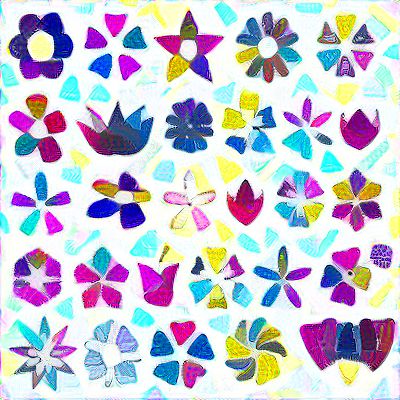}%
    \label{power1}}
    \hfil
    \subfigure[${D^3}$]{\includegraphics[width=0.3\columnwidth]{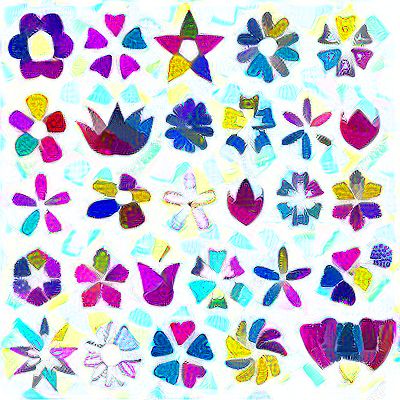}%
    \label{power3}}
    \hfil
    \subfigure[${D^5}$]{\includegraphics[width=0.3\columnwidth]{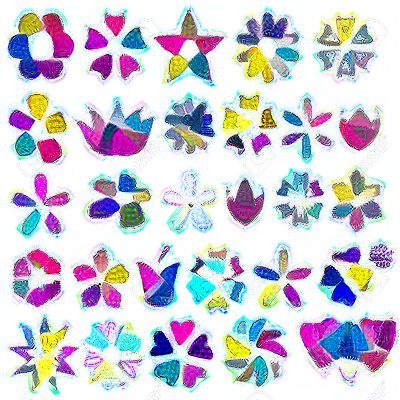}%
    \label{power5}}
  }
  \caption{When there is no emphasis for distance transform image as in \ref{power1}, there are much noise around the original shape.
  With the increase of emphasis power, noise around the shape is removed gradually. (Content image is retrieved from http://clipground.com).}
  \label{emphasis}
\end{figure}

\subsection{The Influence to Neural Style Transfer}

Fig.~\ref{alpha} shows how the distance transform loss ${L_{distance}}$ influences neural style transfer. 
With different values of weighting factors $\alpha$ and $\beta$, different style transfer results can be achieved.
When $\alpha/\beta$ is large, the generated image is more similar to the content image, and when $\alpha/\beta$ is small the generated image is more similar to the style image.

In Fig.~\ref{alpha}, even the $\alpha/\beta$ changed drastically from emphasizing content image to emphasizing style image, there is no change in shapes of silhouettes.
Only inside of shapes are changed according to $\alpha / \beta$ ratio.
Thus, we can assume that although the distance transfer loss ${L_\mathrm{distance}}$ constrains neural style transfer around the silhouette shapes, its influence to style transferring is small.
Because, other than background pixels, distance transfer loss ${L_\mathrm{distance}}$ is too small to be noticed by optimization algorithm.

\subsection{The Strength of Weighting Factor}

The most important aspects of neural style transfer are the weighting factors.
As noted before, different combinations of $\alpha$ and $\beta$ produces different results.
In this experiment, we will the show impact of $\gamma$ with same $\alpha/\beta$.

Fig.~\ref{gamma} shows the results with different values of weighting factor $\gamma$.
Even with small value as in~Fig.\ref{gamma0001}, noises are almost completely removed.
Then when $\gamma$ becomes large enough as in~Fig.\ref{gamma01} and~Fig.~\ref{gamma10}, style transfer is completely contained withing the silhouette shapes. 

\subsection{The Emphasis of Distance Transform Image}

Fig.~\ref{emphasis} shows results with different emphasizing power ${n}$.
When there is no emphasis ( ${n = 1}$ ), there are too much noise formed around the shape. 
When the the emphasizing power ${n}$ is increased, the noises are removed that much. 
Therefore, we can assume that larger emphasizing power is much preferable in order to constrain style transfer around the silhouette shapes.
Because, by taking pixel-wise power of a matrix, large values become much larger and its error will be larger too.
Then, the optimizer tries to remove pixels with large error such as background, resulting in noise free style transfer.

% \begin{figure}%[!t]
% % \centerline{
% % \subfigure[Content Image] {\includegraphics[width=0.3\columnwidth]{squares/humans1.jpg}}%
% % }
% \centerline{
% \subfigure[Large details]{\includegraphics[width=0.3\columnwidth]{squares/big_style.jpg}%
% }
% \hfil
% \subfigure[Medium details]{\includegraphics[width=0.3\columnwidth]{squares/medium_style.jpg}%
% }
% \hfil
% \subfigure[Small details]{\includegraphics[width=0.3\columnwidth]{squares/small_style.jpg}%
% }
% }
% \centerline{
% \subfigure[Generated Image]{\includegraphics[width=0.3\columnwidth]{squares/big.jpg}%
% }
% \hfil
% \subfigure[Generated Image]{\includegraphics[width=0.3\columnwidth]{squares/medium.jpg}%
% }
% \hfil
% \subfigure[Generated Image]{\includegraphics[width=0.3\columnwidth]{squares/small.jpg}%
% }
% }
% \caption{Logo generation using different size and density style images.}
% \label{squares}
% \end{figure}

\begin{figure}%[!t]
\centerline{
\subfigure[Large details]{\includegraphics[width=0.33\columnwidth]{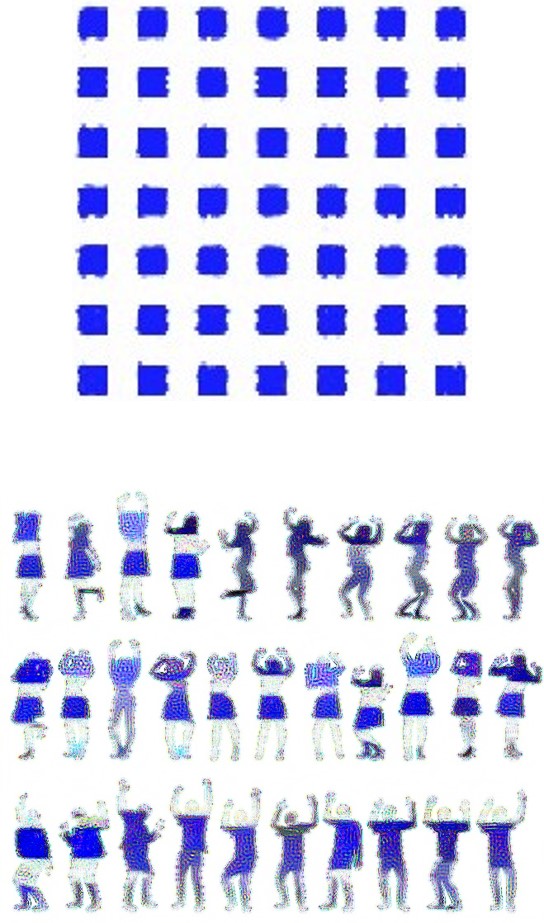}%
}
\hfil
\subfigure[Medium details]{\includegraphics[width=0.33\columnwidth]{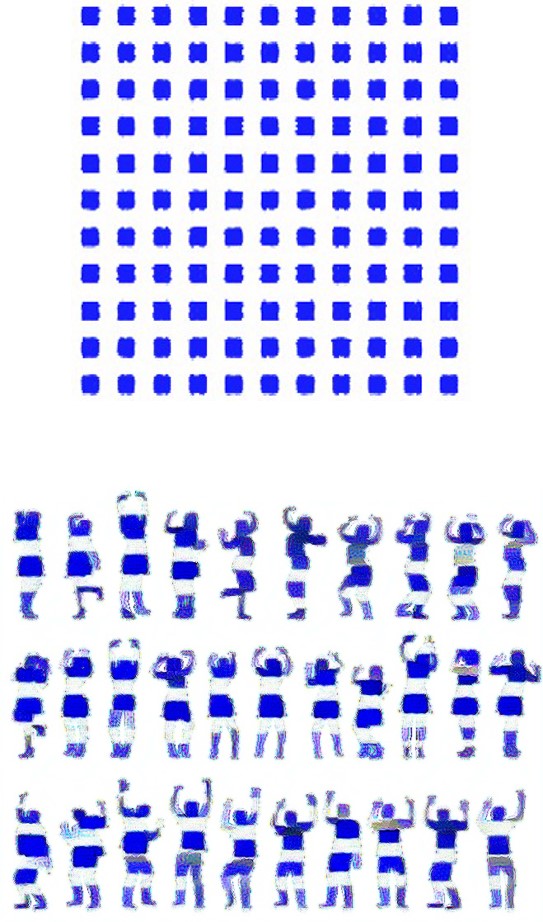}%
}
\hfil
\subfigure[Small details]{\includegraphics[width=0.33\columnwidth]{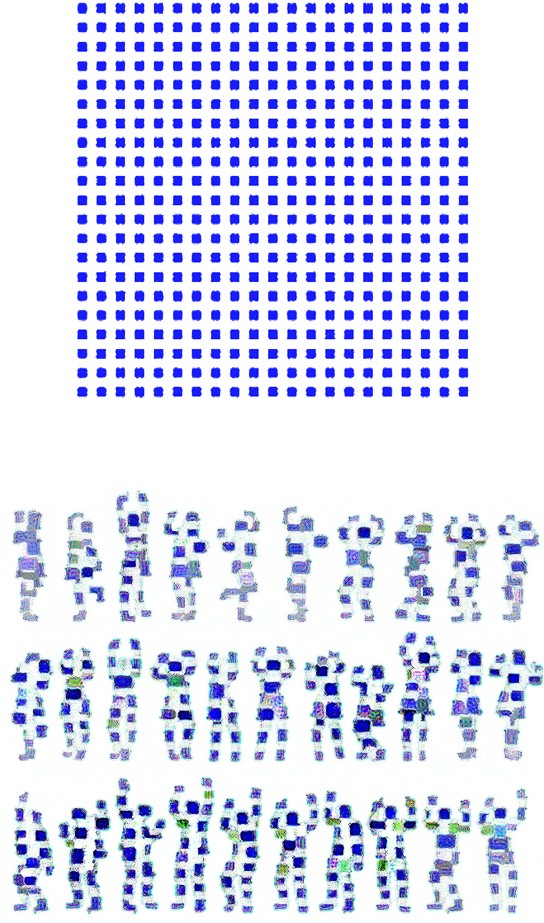}%
}
}
\caption{Logo generation using different size and density style images.}
\label{squares}
\end{figure}

\subsection{Logo Generation}

Fig.~\ref{squares} shows that logos generated using dancing human silhouettes and three different style images with different size and density.
In neural style transfer, size and density of patterns in style image has big impact on resulting image.
When a style image with small details used to synthesize, then the resulting image has small details. 
Whereas, when a style image with big details used, the resulting image has big details.
In addition, density of the style image is transferred to content image too.

For instance in Fig.~\ref{squares}, the human shapes consisted of blue squares of style image in each generated image.
Nevertheless, the size of those squares are different according to the size of details on the style image.
When a style image with small squares is used, the generated image has small squares. 
On the other hand, when a style image with large squares is used, dancing humans consist of large squares in the generated image.

\begin{figure}[!t]
  \centerline{
  	\subfigure[Content Image]{\includegraphics[width=0.3\columnwidth]{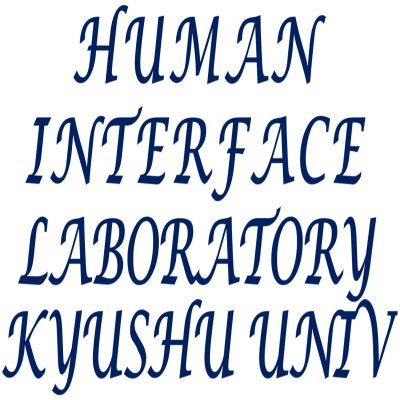}}
    \hfil
    \subfigure[Style Image]{\includegraphics[width=0.3\columnwidth]{fig/colorful_flower.jpg}}
    \hfil
    \subfigure[Generated Image]{\includegraphics[width=0.3\columnwidth]{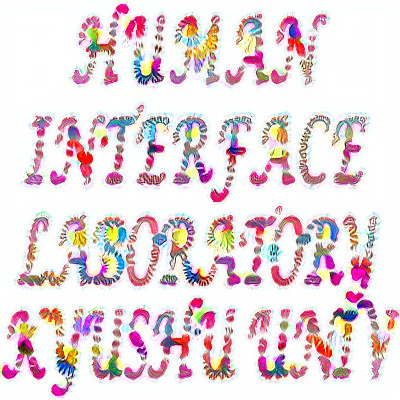}%
    }
  }
  \caption{Logo generation using text image as content image.}
  \label{first_text}
\par\bigskip % force a bit of vertical whitespace
  \centerline{
    \subfigure[Content Image]{\includegraphics[width=0.3\columnwidth]{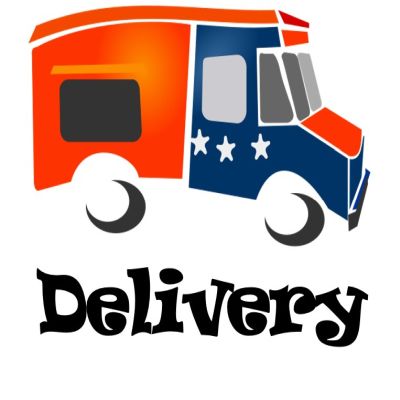}}%
    \hfil
    \subfigure[Style Image]{\includegraphics[width=0.3\columnwidth]{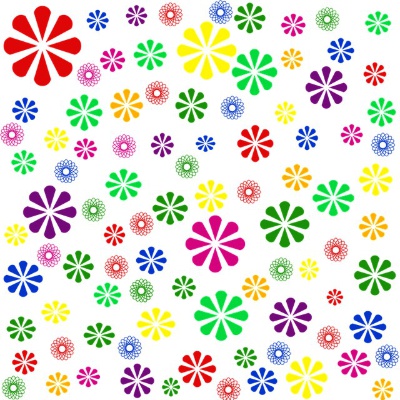}}%
    \hfil
    \subfigure[Generated Image]{\includegraphics[width=0.3\columnwidth]{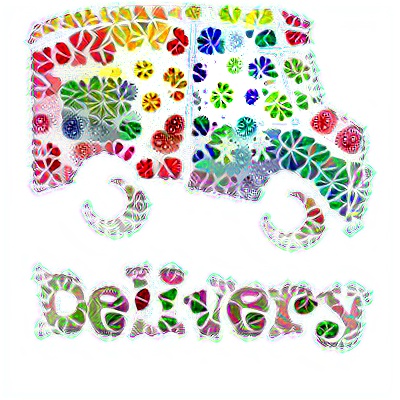}}%
  }
  \caption{Logo generation using shape and text combinations as content image.}
  \label{delivery}
\end{figure}

Fig.\ref{first_text} shows logo generation using a text image as the content image.
Also, some logos have both shapes and text combined.
In Fig.\ref{delivery}, a combination of shapes and text used as the content image for constrained neural style transfer.

\subsection{Transferring Styles to Background}

\begin{figure}[!t]
\centerline{
\subfigure[Content Image]{\includegraphics[width=0.3\columnwidth]{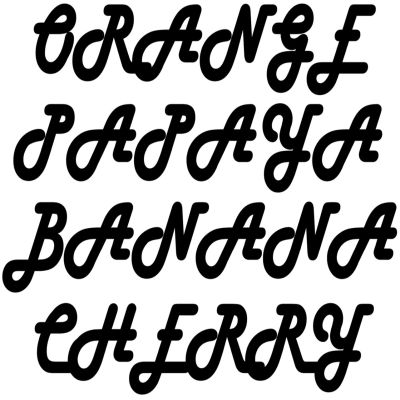}%
}
\hfil
\subfigure[Style Image]{\includegraphics[width=0.3\columnwidth]{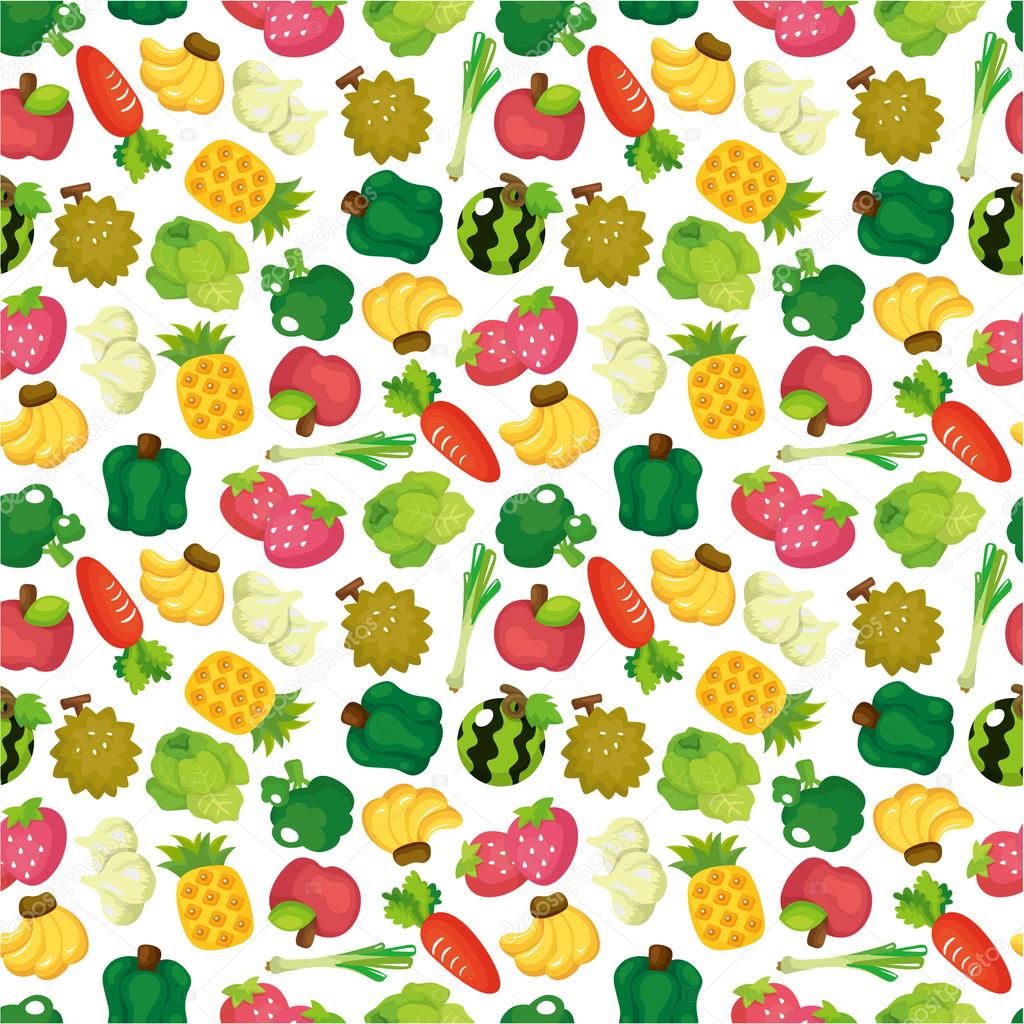}%
}
\hfil
\subfigure[Generated Image]{\includegraphics[width=0.3\columnwidth]{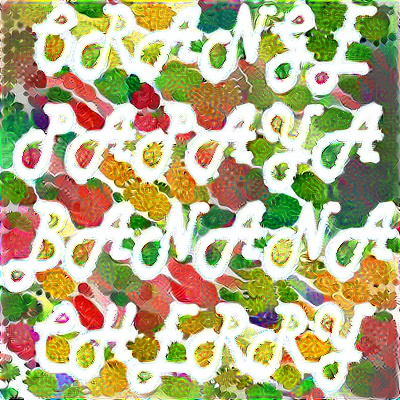}%
}
}
\caption{Logo generation by transferring styles to background.}
\label{background}
\end{figure}

Fig.~\ref{background} shows logo generation by style transferring to the background of a content image.
Looking closer at the generated image, style of the style image appears inside of the shapes.
The reason for that is the shapes in the content image is much thinner and the distance transform values are small.
Thus, errors for pixels inside of shape are small and styles transferred onto them.
To refrain the style from appearing inside of the shapes, much larger emphasizing power and weighting factor are needed.

\subsection{Font Generation}

\begin{figure}[!t]
\centerline{
\subfigure[Content Image]{\includegraphics[width=0.3\columnwidth]{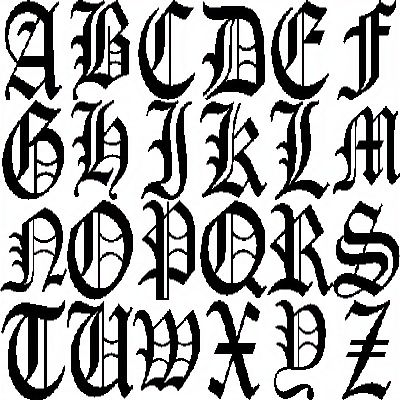}%
\label{good_font}}
\hfil
\subfigure[Bells Style]{\includegraphics[width=0.3\columnwidth]{fig/bells.jpg}%
}
\hfil
\subfigure[Generated Image]{\includegraphics[width=0.3\columnwidth]{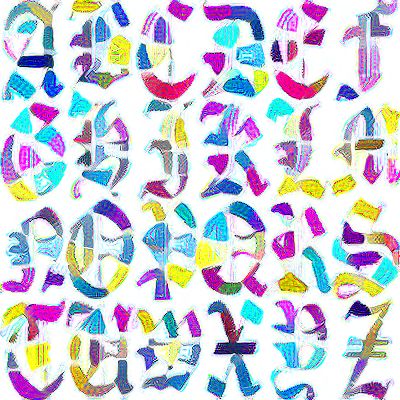}%
}
}
\centerline{
\hspace*{0.325\columnwidth}%
\subfigure[Flowers Style]{\includegraphics[width=0.3\columnwidth]{fig/colorful_flower.jpg}%
}
\hfil
\subfigure[Generated Image]{\includegraphics[width=0.3\columnwidth]{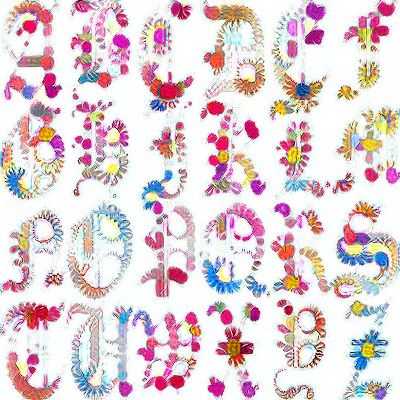}%
}
}
\caption{Font generation using constrained neural style transfer}
\label{make_fonts}
\end{figure}

Using font images as the content image, novel decorated fonts can be generated.
Fig.~\ref{make_fonts} shows that stylized fonts have successfully generated.
Neural style transfer has constrained around the characters due to the distance transform loss.
Also, there is no noise around the characters.

For font generation, it is better to use a style rich content image such as Fig.~\ref{good_font}.
Because, the style transfers onto the characters in the content image, and the characters should be large and wide enough to allow style transferring onto them.
If it was slim font, then styles have nowhere to transfer but very small region of slim characters.

\section{Conclusion}
\label{conclusion}

In this paper, we introduced distance transform loss function for constraining neural style transfer.
While cropping the generated image could be used, it does not produce natural results.
Because the continuity of a generated shape could be cut.
With distance transform loss, neural style transfer occurs in a constrained region, which is determined dynamically.

The proposed method is suitable for using with pre-segmented content images such as silhouette images or clip arts.
We showed that the effect of the distance transform loss is only to constrain style transfer to a region.
It does not interfere with neural style transfer inside of that constrained region.

Also, we observed that how power emphasis and weighting factor work.
Those two parameters should be as large as possible, but if it is too large, it could be same as image cropping.

We generated three different types of logos using three different content images: text images containing only text, shapes such as silhouettes or clip arts, and images that contain shapes and text both.
% In all three cases constrained neural style transfer had achieved an interesting results.
In addition to that, we demonstrated logo generation by transferring to the background of a content image.
In this case, parameters should be tuned differently to achieve successful results. 
The biggest advantage of the logos generated by constrained neural style transfer, is that generated logos are genuine and novel without any fear of a duplicate.

Finally, we provided examples of font generation with constrained neural style transfer.
Completely novel and clean font has been generated by out method.
In the future research, we are planning to build a complete system for generating logos.
%However, one of the downsides of proposed method is content images and style images have to be provided by the user.
%For perfect logo generation by synthesis, good content image and style image combination should be provided.
%Also, we didn't consider about the psychology and meaning of logos.
%In the future research, we are planning to address above problems, as well as building a complete system for generating logos.

\section*{Acknowledgement}
This research was partially supported by MEXT-Japan (Grant No.J17H06100).

\bibliographystyle{IEEEtran}
\bibliography{styletransfer}

\end{document}